\documentclass{article}

\usepackage[preprint]{neurips_2026}

\usepackage[utf8]{inputenc} 
\usepackage[T1]{fontenc}    
\usepackage{hyperref}       
\usepackage{url}            
\usepackage{booktabs}       
\usepackage{amsfonts}       
\usepackage{nicefrac}       
\usepackage{microtype}      
\usepackage{xcolor}         
\usepackage{graphicx}
\usepackage{amsmath}
\usepackage{amssymb}
\title{AlterNet: Adaptive-Depth Mixture-of-Experts with Learned Halting for Elastic Inference}

\author{%
  Alexander Kastius \\
  SAP\\
  Palo Alto, CA 94304 \\
  \texttt{a.kastius@sap.com} \\
}

\begin{document}

\maketitle

\begin{abstract}
  Most LLM architectures have fixed compute demands determined at training time. We propose a novel architecture that combines the Mixture-of-Experts architecture with shallow experts and a novel dynamic-depth system to make the extent of a forward pass dynamically adjustable. Our approach lets the model automatically determine the compute needed for a given task, paired with a system to scale it down after training without retraining when the execution environment demands faster or cheaper inference. To achieve this, we introduce an iterative computation mechanism, sequence-level depth-based routing steered by a halting mechanism within the experts, a depth encoding that lets the model track its progress in the chain, and a regularization mechanism that incentivizes the model to use allocated compute efficiently. We present initial benchmarks and discuss future research directions.
\end{abstract}

\section{Introduction}
Adapting the computational demands of large models to resource-constrained environments is a heavily studied research area, cf. \cite{DBLP:journals/csur/Menghani23}. Proposed solutions include quantization \citep{DBLP:journals/corr/abs-2208-07339}, distillation \citep{DBLP:journals/corr/HintonVD15}, and others. In many cases, for example, in the Qwen model line \citep{DBLP:journals/corr/abs-2412-15115}, scaled-down versions of the models with either lower precision or fewer parameters are presented. In those cases, it is still necessary to retrain or at least partially reparametrize a model. In many cases, the model's capacity is also reduced across all domains, unless it is specially retrained.

One architecture option in particular supports those efforts further: By now, Mixture-of-Experts (MoE) models are commonly observed among releases, cf. \citet{DBLP:journals/jmlr/FedusZS22}, \citet{DBLP:journals/corr/abs-2401-04088}. MoE distinguishes between active and passive parameters, contained in active and passive experts, with the latter remaining unused. A central router model decides which experts stay active. This allows using only a fraction of the model, at the cost of slightly lower performance. MoE models have their own disadvantages, most importantly that experts are usually disconnected to some extent; i.e., the knowledge of one expert does not become available as input to other experts.

We propose a design concept to overcome these limitations. If the intuition about the underlying mechanism of large experts holds, knowledge is distributed topic-wise among them. If so, interchanging information between them to allow further specialization can be beneficial. We propose a mechanism in which information flows between alternating experts across iterations, which we call Alternet. Combined with an explicit depth signal, this allows experts to specialize for distinct positions within the loop.
Furthermore, not all problems require the same computational depth. Following from this, we propose a mechanism to express an expert's belief in the quality of its output, allowing early stopping when sufficient confidence is observed. The per-expert halt heads and the sinusoidal depth encoding are our two most relevant additions to previous iterations on the matter.

\section{Related Work}
\label{sec:relatedwork}
Several research directions relate to our work, including sparse and iterative architectures, as well as other mechanisms to dynamically adjust compute. \citet{DBLP:journals/corr/Graves16} present one of the earliest available examples of a mechanism that implements an adaptive halting mechanism for recurrent neural networks (RNNs), which highly influenced the second important approach, PonderNet \citep{DBLP:journals/corr/abs-2107-05407}. PonderNet provides the mechanism we later reuse. We vastly extended the fundamental idea by applying it to a deep Mixture-of-Experts transformer architecture, replacing the prior function and introducing distributed halt heads.

Several other projects have already evaluated iterative or recurrent transformers, most notably \citet{DBLP:conf/iclr/DehghaniGVUK19}. Universal Transformers combines self-attention with a recurrent loop and ACT halting, a predecessor of the halting mechanism used by PonderNet. \citet{DBLP:journals/corr/abs-2606-18206} present another looped approach that stops dynamically when the model's internal state becomes stable, while relying on large iteration counts. \citet{DBLP:journals/corr/abs-2510-25741} is one of the closest competitors to this method; instead of an MoE architecture, a single-block transformer is used, and the halt signal is based on per-step improvements, a method we adopt. Another PonderNet successor, cf. \citet{DBLP:journals/corr/abs-2603-01914}, applies the halting setup on a per-token level using a dense model and global depth heads.

MoE approaches have reshaped the landscape for sparse models, cf. \citet{DBLP:conf/iclr/ShazeerMMDLHD17}, which introduced sparse routing across many small networks, \citet{DBLP:journals/jmlr/FedusZS22}, using simpler top-1 routing and the router balancing loss, and \citet{zoph2022stmoedesigningstabletransferable}, introducing the z-loss for the router used in this evaluation. \citet{DBLP:journals/corr/abs-2605-09165} shows MoE can help overcome the critical challenges of looped setups. They do not implement adaptive halting. Furthermore, \citet{DBLP:journals/corr/abs-2606-04438} displays another looped approach, which again uses a fixed number of iterations. Depth encoding is used only by \citet{DBLP:journals/corr/abs-2606-04438}, where it is injected into the state instead of being explicitly passed to the relevant components. An alternative, but orthogonal approach is shown using Mixture-of-Depths, c.f. \citet{DBLP:journals/corr/abs-2404-02258}, which lets certain parts of the input skip further computation.

\section{Architecture}
\label{sec:architecture}
\begin{figure}[t]
  \centering
  {\fontsize{6}{7}\selectfont
  \includegraphics[width=0.6\linewidth]{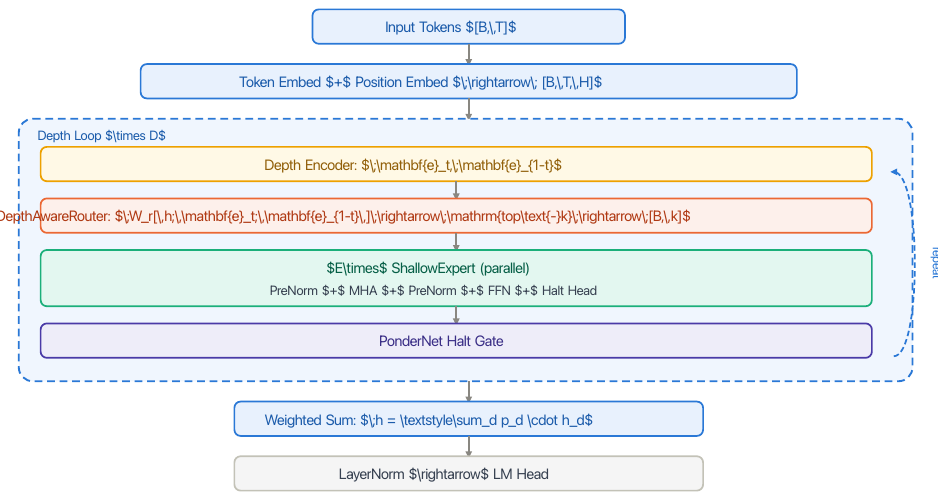}}
\vspace{-0.7cm}
  \caption{AlterNet forward pass. Input tokens are embedded and passed through a depth loop with a maximum of $D$ iterations. At each step, a depth-aware router selects $k$ of $E$ shallow experts; a PonderNet halt gate accumulates per-step probabilities against a Gaussian prior.}
  \label{fig:architecture}
\end{figure}

AlterNet adds several components to enable distributed, scalable computation and downscaling without retraining. Figure \ref{fig:architecture} illustrates the proposed setup. Our system resembles a conventional MoE setup, with several important changes. Generally, information flows from the input to the router to the selected shallow experts. If the halting mechanism concludes that further inference is necessary, the process repeats.

The input tokens are projected into the initial hidden state using learned position and token embeddings. Only at this step is horizontal positional information injected into the token stream. Later iterations do not re-inject horizontal information, leaving the internal state uncorrupted.

The first step within each iteration computes the depth encoding. The depth encoding is the vertical component of the positional encoding. It gives the router a way to evaluate whether a more specialized expert suited to later iterations should be considered. The signal is based on the iteration index $t$, with $d$ being the current iteration, starting from $0$, and $D$ being the maximum allowed iteration at training time, a highly relevant hyperparameter:

\vspace{-0.3cm}
\begin{equation}
    t = \frac{d}{D-1}
\end{equation}
\vspace{-0.3cm}

The resulting embedding is computed using the same sinusoidal encoding used in classic transformer architectures \citep{DBLP:conf/nips/VaswaniSPUJGKP17}. We decided against RoPE, cf. \citet{DBLP:journals/ijon/SuALPBL24}, to improve generalization to OOD depths. Two depth signals are passed to the router: $\sin(t)$ and $\sin(1-t)$, with the latter providing explicit information about the remaining compute budget. It is intentionally not added to the residual stream to avoid accumulation in the state. Two 64-d sinusoidal embeddings $\mathbf{e}_t$, $\mathbf{e}_{1-t}$ are concatenated to the router input; being parameter-free, they extrapolate gracefully beyond training depth. The depth-aware router consists of a single linear layer with weight matrix $W_r \in \mathbb{R}^{E \times (H + 128)}$, relying on the input state at the per-sequence, not per-token, level, as well as the two depth-encoding signals. Let $x_d^{(b)} \in \mathbb{R}^{T \times H}$ denote the hidden state for sequence $b$ at depth step $d$, with $x_0^{(b)}$ being the token+position embedding output. This is a fundamentally different approach from most MoE setups today; cf. 
\citet{DBLP:journals/jmlr/FedusZS22} and \citet{DBLP:journals/corr/abs-2401-04088}. We use the last valid token in the hidden state as an aggregator for the whole sequence, reducing routing overhead and eliminating token-level variance as a cause of router instability. Given sequence aggregator $h_d^{(b)} = x_d^{(b)}[T,:]$ and logits $\mathbf{l}_d^{(b)} = W_r[h_d^{(b)};\,\mathbf{e}_t;\,\mathbf{e}_{1-t}]$, the gate weights are top-$k$ softmax probabilities:
\begin{equation}
    g_j^{(b)} = \frac{\exp(l_{d,j}^{(b)})}{\sum_{i \in \mathrm{top\text{-}k}(b)} \exp(l_{d,i}^{(b)})}, \qquad
    \mathrm{top\text{-}k}(b) = \operatorname{argtopk}_{k}\!\left(\mathbf{l}_d^{(b)}\right)
\end{equation}

We use the $z$-loss $\mathcal{L}_{z} = \mathbb{E}\!\left[\left(\log Z\right)^{2}\right]$, introduced in \citet{zoph2022stmoedesigningstabletransferable}, to ensure router stability, with $Z = \sum_i e^{l_{d,i}^{(b)}}$. The logits are furthermore clamped to $10$. The load-balance auxiliary loss follows \citet{DBLP:journals/jmlr/FedusZS22}. $\mathcal{L}_{\mathrm{LB}} = E \cdot \sum_{i=1}^{E} f_i \cdot P_i$, where $E$ is the number of experts, $f_i$ is the fraction of examples routed to expert $i$ in the batch, and $P_i$ is the mean softmax routing probability assigned to expert $i$. Minimizing this loss encourages equal utilization across experts. The set of experts follows the router. As looping allows depth, we kept the experts themselves shallow, with each consisting of a single transformer block. Each expert's internal layout is a chain of multi-head attention to a two-layer feed-forward network (FFN). Each expert has one additional scalar output, the halt-head $\hat{\lambda}_k$, generated by a linear layer with weights $\mathbf{w}_k$ and bias $b_k$, with the latter initialized to $-3.0$. It receives the output of the expert $y_k^{(b)} = \mathrm{expert}_k(x_d^{(b)})$ as input, and again relies on the last valid token as aggregation: $\hat{\lambda}_k^{(b)} = \mathbf{w}_k^\top y_k^{(b)}[T, :] + b_k$. Our goal is for the experts to use this to encode confidence in whether their output should represent the network's final state.

During training, the central loop is limited to a preset maximum number of iterations, $D$, as used in the depth encoding. Within the loop, the encoding is computed, $k$ experts are selected, and their output is generated. The outputs are added to the residual stream weighted by their initial gate weights, and the halt heads are piped into the halting mechanism. Given the full hidden state over all tokens of shape $[B{\times}T{\times}H]$, $x_d^{(b)}$ and each experts output $y_j^{(b)}$, the output of each iteration becomes:

\begin{equation}
    x_{d+1}^{(b)} = x_d^{(b)} + \sum_j g_j^{(b)} \cdot y_j^{(b)}
\end{equation}

The halting mechanism follows the training setup proposed in PonderNet \citep{DBLP:journals/corr/abs-2107-05407}. The key addition is incorporating expert-level halt heads $\lambda_d^{(b)}$ to improve the granularity of the halt signal. With $d$ being the current depth level, the halting probability is denoted as $p_d^{(b)}$. Given $g^{(b)}_k$ as the router gate weight of each expert, $p_d^{(b)}$ becomes:

\vspace{-0.2cm}
\begin{align}
    \lambda_d^{(b)} &= \sum_{j \in \mathrm{top\text{-}k}(b)} g_j^{(b)} \cdot \sigma(\hat{\lambda}^{(b)}_j)
    p_d^{(b)} &= p_{\mathrm{unhalted},d}^{(b)} \cdot \lambda_d^{(b)}, \qquad
p_{\mathrm{unhalted},d+1}^{(b)} = p_{\mathrm{unhalted},d}^{(b)} \cdot \left(1 - \lambda_d^{(b)}\right)\\
\end{align}

\vspace{-0.2cm}
This induces a probability distribution over iterations, shaped by the weighted halt heads of the selected experts, which is then KL-regularized to match a prior distribution modeling the expected number of iterations. PonderNet relied on a geometric distribution. We argue that a geometric distribution does not correctly model the task landscape the model faces. If the maximum depth is correctly tied to the complexity of the tasks the model is actually trying to represent, a natural interpretation is that the capacity required for randomly sampled tasks forms a bell shape, rather than being skewed toward the lower end. Following this hypothesis, we use a Gaussian distribution with a mean centered around half of the maximum number of iterations: $q_d \propto \mathcal{N}(d;\;\mu{=}D/2,\;\sigma{=}D/4)$.

Later experiments further motivate this choice, as all experiments using the geometric prior were unsuccessful, as it further supported the depth-bootstrap collapse problem with a high probability mass being assigned to depth-1; see Section \ref{sec:benchmark} for a detailed description of our observations.

The system contains three learning entities: The router, the halt heads, and the actual experts. All three must be tuned to achieve their corresponding goals, leading to a comparatively complex loss landscape. Overall, the loss function has five major components. Using $\mathcal{L}_{\mathrm{LB}}$ and $\mathcal{L}_{z}$ as defined above and $\mathcal{L}_{\mathrm{CE}}(x_d^{(b)})$ as the per-step cross-entropy, the full learning objective becomes:

\begin{align}
    \mathcal{L} &= \mathcal{L}_{\mathrm{PN}} + \alpha_{\mathrm{KL}} \cdot \mathcal{L}_{\mathrm{KL}} + \alpha_{\mathrm{LB}} \cdot \mathcal{L}_{\mathrm{LB}} + \alpha_{z} \cdot \mathcal{L}_{z} + \alpha_{\mathrm{RIR}} \cdot \mathcal{L}_{\mathrm{RIR}} \\
    \mathcal{L}_{\mathrm{PN}}  &= \frac{1}{B}\sum_{b=1}^{B} \sum_{d=0}^{D-1} p_d^{(b)} \cdot \mathcal{L}_{\mathrm{CE}}(x_d^{(b)}) \\
    \mathcal{L}_{\mathrm{KL}}  &= \frac{1}{B}\sum_b \sum_{d=0}^{D-1} p_d^{(b)} \log \frac{p_d^{(b)}}{q_d} \\
    \mathcal{L}_{\mathrm{RIR}} &= \frac{1}{B}\sum_{b=1}^{B}\frac{1}{D-1} \sum_{d=1}^{D-1} \mathrm{BCE}\!\left(\lambda_d^{(b)},\; 1 - w_d^{(b)}\right) \label{eq:rir}
\end{align}

$\mathcal{L}_{\mathrm{PN}}$ is the PonderNet cross-entropy, measuring the summed per-step cross-entropy of each iteration's output. $\mathcal{L}_{\mathrm{KL}}$ is the Kullback-Leibler divergence measured in relation to the Gaussian prior. $\mathcal{L}_{\mathrm{RIR}}$ is the additional loss used to measure per-iteration improvement and further incentivize the network to stop as intended. It follows the loss function specified in Equation \ref{eq:rir}, with $w_d^{(b)}$ being defined as:

\vspace{-0.3cm}
\begin{equation}
w_d^{(b)} = \sigma\!\left(50 \cdot (\max(\mathcal{L}_{\mathrm{CE}}(x_{d-1}^{(b)}) - \mathcal{L}_{\mathrm{CE}}(x_d^{(b)}),0) - 0.005)\right)
\end{equation}
\vspace{-0.3cm}

It measures the gradual improvement at each iteration and incentivizes the halt heads to stop when the improvement becomes small. This aspect of the loss is detached from the rest of the graph and affects only the halt-head weights. The challenge in this environment is balancing these factors correctly.

\section{Evaluation}
\label{sec:benchmark}
In our evaluation, we initially ran ablations across several dimensions to assess our setup's sensitivity to the new hyperparameters. The ablations are described in detail in Appendix \ref{app:ablations}. Three main conclusions from the ablations guided the following experiments: High $\alpha_{\mathrm{KL}}$ values are required to actually guide the model towards dynamic depth, the budget signal has a positive effect, removing it hurts training, and the per-expert halt heads perform better than a central halt head.

To provide a baseline, we measured the token perplexity of OPT 350M on the C4 dataset, cf. \citet{DBLP:journals/corr/abs-2205-01068}. When evaluated, the model yields a perplexity of approximately 21-22. Compared to our experiments, training the OPT models consumed significantly more data, up to 180B tokens. Until we can complete training on a full-scale model, we omit further performance indicators. We trained two models with 350M parameters. The setup is outlined in Appendix \ref{app:hyperparams}. The experiments consisted of: I, a setup in which we fixed the model to twelve iterations to serve as a baseline, and II, a setup in which we allowed dynamic compute allocation. All models have been significantly undertrained, cf. \cite{DBLP:journals/corr/abs-2203-15556}, due to time constraints. The 30\,000 step run consumed about 3.9B tokens, well below the suggested 7B tokens. For all but model II, we stopped training after 10K steps because evaluation performance did not continue to grow enough to justify the required compute budget. For the dynamic halting setup, we continued training for another 20K steps, as the performance increase between evaluation points was still significant.

A key finding from our 350M experiments is that we must overcome the depth-bootstrap collapse problem: depth 0 is the only depth level whose loss weight is independent of other halting probabilities. All deeper levels receive weight $p_d = \prod_{i<d}(1-\lambda_i)\cdot\lambda_d$, which collapses toward zero once the depth-0 head grows confident. This leads to faster convergence for the depth-1 cross-entropy, which in turn leads the model to not use later layers at all. A geometric prior cooperates with the bootstrap collapse. We introduce an aligned curriculum release for training, during which the model is forced to stick to its maximum number of iterations, and the Gaussian-centered prior as a practical solution. Three intermediate runs on the 350M setup (geometric prior, slower $\alpha_{\mathrm{KL}}$ ramp under geometric and curriculum alone) each collapsed or failed to generalize beyond depth 1; the combination was the only one to remain stable, cf. Appendix \ref{app:collapse}.

\begin{figure}[t]
  \begin{minipage}[t]{0.49\linewidth}
    \centering
    \includegraphics[width=\linewidth]{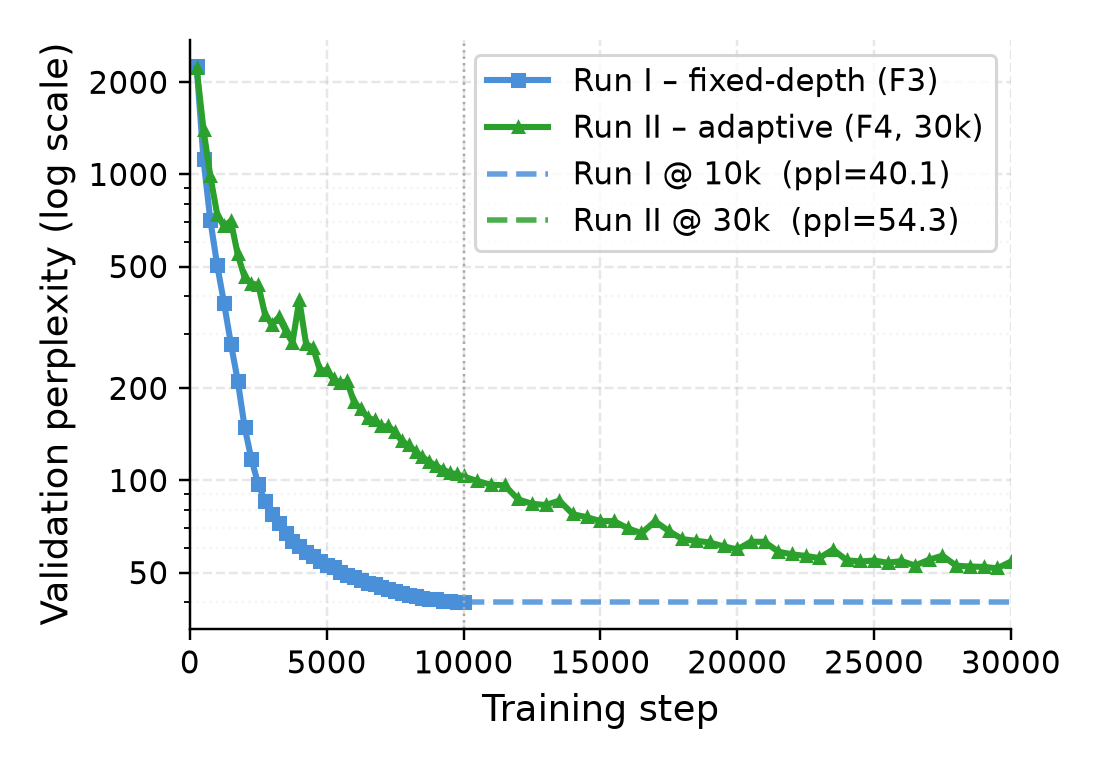}
    \caption{Evaluation performance. Model I is fixed at 12 iterations; Model II adapts dynamically.}
    \label{fig:ppl}
  \end{minipage}
  \hfill
  \begin{minipage}[t]{0.49\linewidth}
    \centering
    \includegraphics[width=\linewidth]{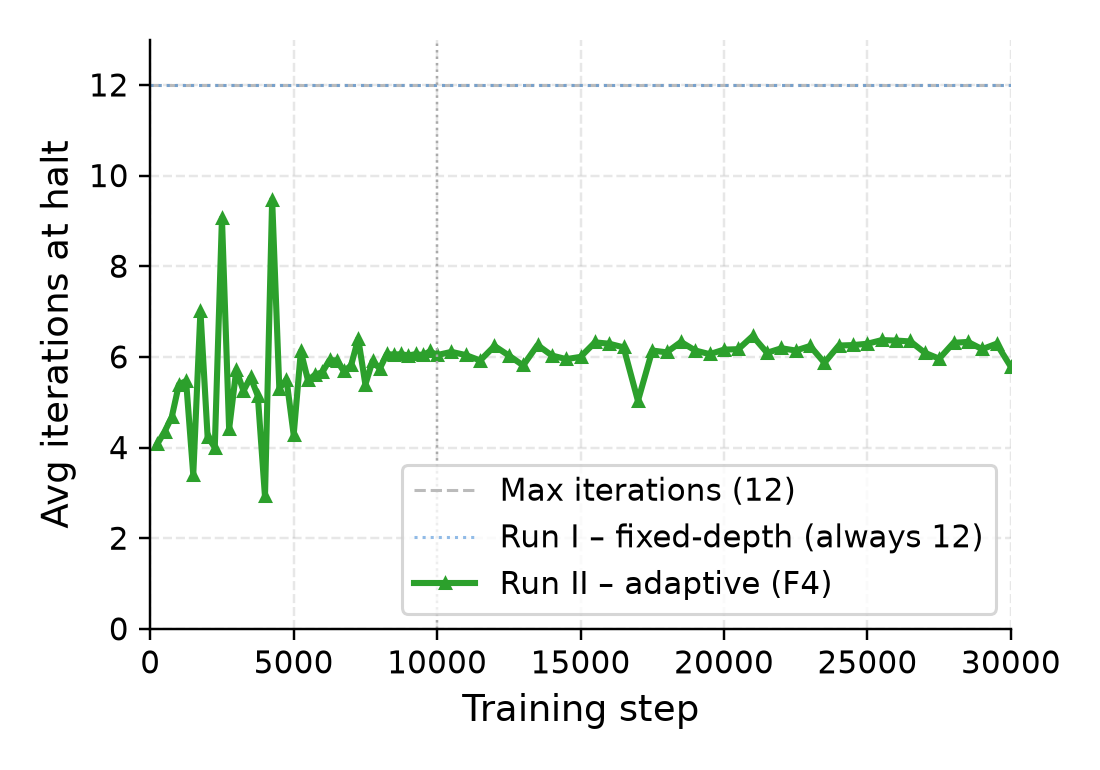}
    \caption{Average iterations per forward pass during training. The model converges toward the Gaussian prior, saving $\approx$50\% of iterations.}
    \label{fig:iters}
  \end{minipage}
  \vspace{-0.2cm}
\end{figure}

The evaluation yielded several results worth mentioning, illustrated by Figures \ref{fig:ppl} and \ref{fig:iters}: Without further optimization, under the given training constraints, we did not yet reach the performance of comparable models with our setup, either with Model I or II.  Furthermore, Model II yields a 54.3 PPL compared to the 40.1 PPL achieved by Model I at 48.3\% of the required number of iterations during inference. The savings of approximately 6 iterations can be realized once training is finished. To draw final conclusions, continuing training to observe sufficient tokens relative to the number of parameters would be required. At equal step counts, Model II lags Model I in evaluation performance due to the overhead of tuning the halting mechanism. After training for approximately three times as many tokens, the equally sized dynamic halting model reaches a comparable order of magnitude.

\section{Outlook}
\label{sec:conclusions}
We have shown that scaling the training mechanism to achieve inference performance gains is possible, albeit at a price. While we overcome several central limitations, especially depth 1 collapse, we leave several relevant questions unanswered. We aim to answer those in a future manuscript. At this point, continuing model training with the assumed optimal number of tokens and beyond is a prerequisite for concluding the model's maximum performance. Perplexity alone is not sufficient; it is merely an indicator of potential. All our evaluations stopped early and were run on comparatively small models. Scaling up, in terms of absolute model size, maximum allowed iteration depth, and scaling out, in terms of expert network width and router capacity, are two required steps. Furthermore, our setup allows scaling down the model during inference to adapt to environments with limited compute budgets or time constraints; we will follow with a precise analysis of this behavior. Furthermore, during inference it might be possible to intelligently drop or preload rarely used experts when the model is used in a certain domain based on depth, an additional improvement we consider worth evaluating under tight memory constraints.

\bibliographystyle{plainnat}
\bibliography{references}

\newpage
\appendix

\section{Hyperparameters and Experimental Setup}
\label{app:hyperparams}

\subsection*{Model Hyperparameters}
\begin{table}[h]
\centering
\caption{AlterNet model hyperparameters for ablation (50M) and scale-up (350M) experiments.}
\begin{tabular}{lcc}
\toprule
Hyperparameter & 50M (ablations) & 350M (scale-up) \\
\midrule
Hidden size $H$ & 512 & 2048 \\
Number of experts $E$ & 8 & 10 \\
Top-$k$ & 2 & 2 \\
FFN dimension per expert & 2048 & 2048 \\
Max iterations $D$ & 8 & 12 \\
Attention heads & 8 & 16 \\
Depth embedding dim & 32 & 64 \\
Vocabulary size & 50{,}257 & 50{,}257 \\
Sequence length & 512 & 512 \\
Halt head bias init & $-3.0$ & $-3.0$ \\
Total parameters & $\approx$50M & $\approx$350M \\
\bottomrule
\end{tabular}
\end{table}

\subsection*{Training Hyperparameters}
\begin{table}[h]
\centering
\caption{Training setup for ablation and scale-up versions}
\begin{tabular}{lcc}
\toprule
Hyperparameter & 50M ablations & 350M Run I / Run II \\
\midrule
Effective batch size & 64 & 256 \\
Tokens per step & 32{,}768 & 131{,}072 \\
Learning rate & $3\times10^{-4}$ & $3\times10^{-4}$ / $3\rightarrow1\times10^{-4}$ \\
LR warmup steps & 1{,}000 & 2{,}000 \\
Training steps & 5{,}000 & 10{,}000 / 30{,}000 \\
Total tokens seen & $\approx$164M & $\approx$1.3B / 3.9B \\
Precision & BF16 & BF16 \\
$\alpha_z$ (z-loss weight) & $10^{-3}$ & $10^{-3}$ \\
$\alpha_{\mathrm{LB}}$ (load-balance weight) & $10^{-2}$ & $10^{-2}$ \\
$\max \alpha_{\mathrm{KL}}$ (KL weight) & $0.1$ & $1.0$ (ramped) \\
$\alpha_{\mathrm{KL}}$ warmup steps & 1{,}000 & 1{,}000 \\
Min-depth warmup steps & -- & 1{,}000 (Run II) \\
Prior type & Geometric ($p=0.12$) & Gaussian ($\mu=6, \sigma=3$) (Run II) \\
$\alpha_{\mathrm{RIR}}$ & 0 & 0.2 (Run II) \\
Dataset & C4 (en) & C4 (en) \\
\bottomrule
\end{tabular}
\end{table}

\section{Full Ablation Results}
\label{app:ablations}

All ablations were executed using small models with approximately 50M parameters to gain a basic understanding of the system mechanics. The following experiments were run:

\begin{itemize}
    \item \textbf{Group A, expert size:} We evaluated four setups with $E$ experts and a $d_{\operatorname{ffn}}$ feedforward dimension: $(E=8,d_{\operatorname{ffn}}=2048)$, $(E=16,d_{\operatorname{ffn}}=1024)$,$(E=32,d_{\operatorname{ffn}}=512)$,$(E=64,d_{\operatorname{ffn}}=256)$.
    \item \textbf{Group B, budget signal:} The budget signal in the depth encoding tends to improve performance on a small scale. Dropping it increased the perplexity after 5000 steps by 20 points.
    \item \textbf{Group C, expert halt heads:} Replacing the expert-level halt heads with a global halt head hurts performance by 11 PPL.
    \item \textbf{Group D, $\alpha_{\mathrm{KL}}$-evaluation:} $\alpha_{\mathrm{KL}}$ needs to be set to adequately high values, 0.1 on 50M, and at least 1.0 on 350M, to actually yield a model that produces a diverse number of iterations. We evaluated $\alpha_{\mathrm{KL}} \in \{0.001, 0.01, 0.1, 1.0\}$.
    \item \textbf{Group E, iteration value:} Each model size and setup has a performance peak at a certain number of iterations. Raising the number of allowed iterations from smaller values towards 12 consistently improved performance. Further increases did actually hurt the short-term performance during training. We do not expect this to be strictly transferable to scaled-up setups.
    \item \textbf{Group F, other parameters:} Neither the z-loss weight nor the logit clamp has a significant impact on this scale.
\end{itemize}

A more detailed listing of the results follows. Based on those results, we ran the final evaluation. All ablations were executed using small models and small datasets; we do not claim that the conclusions drawn above hold at any scale. Furthermore, the experiments are single-seed estimates. All ablation runs use the 50M preset, ran over 5{,}000 steps with an effective batch size of 64. The most fundamental baseline established was A0 stopping at a perplexity of 1342 within the short training run, but sticking to fixed 3 iterations. Most ablations showed this behavior of sticking to a specific depth; only raising the weight of the KL divergence lifted this force.

\subsection*{Group A --- Expert Granularity}
Top-$k=2$ throughout; only expert count and FFN dimension vary.
\begin{table}[h]\centering
\begin{tabular}{lrrrrrrr}
\toprule
Run & Experts & ffn\_dim & ppl & lm\_loss & iters & savings & cv \\
\midrule
A0/G0 (baseline) & 8 & 2048 & 1342 & 7.202 & 3.00$\pm$0.01 & 62.5\% & 0.31 \\
A1-tk2 (clean) & 16 & 1024 & 1394 & 7.240 & 2.00$\pm$0.00 & 75.0\% & 0.56 \\
A2 & 32 & 512 & 1437 & 7.270 & 3.00$\pm$0.03 & 62.5\% & 0.85 \\
A3 & 64 & 256 & 1497 & 7.311 & 3.00$\pm$0.08 & 62.5\% & 1.65 \\
\bottomrule
\end{tabular}
\caption{Monotonic degradation as expert count increases and ffn\_dim decreases. Experts below ffn\_dim$\approx$1024 lack sufficient capacity to specialize.}
\end{table}

\subsection*{Group B --- Depth and Budget Signal Ablation}
\begin{table}[h]\centering
\begin{tabular}{lllrrrr}
\toprule
Run & Depth sig. & Budget sig. & ppl & iters & savings & cv \\
\midrule
B1 & \checkmark & -- & 1362 & 3.00$\pm$0.00 & 62.5\% & 0.21 \\
B3/G0 (default) & \checkmark & \checkmark & 1342 & 3.00$\pm$0.01 & 62.5\% & 0.31 \\
\bottomrule
\end{tabular}
\caption{Removing the sinusoidal budget signal $\sin(1-t)$ increases ppl by 20 points ($+1.5\%$). B0 (no depth signal at all) and B2 (budget-only) remain untested.}
\end{table}

\subsection*{Group C --- Halt Mechanism}
\begin{table}[h]\centering
\begin{tabular}{llrrr}
\toprule
Run & Halt type & ppl & iters & savings \\
\midrule
C0 & Global single head & 1354 & 2.00$\pm$0.00 & 75.0\% \\
C1 (default) & Per-expert (aggregated) & 1343 & 3.00$\pm$0.00 & 62.5\% \\
\bottomrule
\end{tabular}
\caption{Per-expert halt outperforms global halt by 11 ppl. Both lock to a fixed depth.}
\end{table}
\newpage
\subsection*{Group D --- KL Weight ($\alpha_{\mathrm{KL}}$) Sweep}
\begin{table}[h]\centering
\begin{tabular}{lrrrrrr}
\toprule
Run & $\alpha_{\mathrm{KL}}$ & ppl & lm\_loss & iters & savings & cv \\
\midrule
D1 & 0.001 & 1347 & 7.206 & 3.00$\pm$0.00 & 62.5\% & 0.18 \\
D2/G0 (default) & 0.01 & 1342 & 7.202 & 3.00$\pm$0.01 & 62.5\% & 0.31 \\
D3 & 0.1 & 1351 & 7.209 & 2.65$\pm$0.46 & 66.9\% & 0.25 \\
D4 & 1.0 & 1418 & 7.257 & 3.37$\pm$1.83 & 57.9\% & 1.04 \\
\bottomrule
\end{tabular}
\caption{$\alpha_{\mathrm{KL}}=0.1$ achieves the only genuine per-example depth variance (std=0.46) with negligible quality cost ($+9$ ppl). $\alpha_{\mathrm{KL}}=1.0$ overwhelms the task loss on very small models.}
\end{table}

\subsection*{Group E --- Iteration Ceiling Sweep}
\begin{table}[h]\centering
\begin{tabular}{lrrrrrr}
\toprule
Run & max\_iter & ppl & iters & savings & cv & Notes \\
\midrule
E1 & 4 & 1383 & 4.00$\pm$0.00 & 0.0\% & 0.81 & Never halts early \\
E2/G0 (default) & 8 & 1342 & 3.00$\pm$0.01 & 62.5\% & 0.31 & \\
E3 & 12 & 1334 & 3.00$\pm$0.00 & 75.0\% & 0.22 & Best quality \\
E4 & 16 & 1358 & 3.00$\pm$0.00 & 81.3\% & 0.35 & \\
E5 & 24 & 1389 & 3.00$\pm$0.00 & 87.5\% & 0.83 & Routing degrades \\
E6 & 32 & \multicolumn{5}{l}{INVALID ($\times$2) --- KL divergence at step 1; requires longer $\alpha_{\mathrm{KL}}$-warmup} \\
\bottomrule
\end{tabular}
\caption{Halt depth locks at 3 regardless of ceiling. Quality peaks at iter=12 and degrades for higher ceilings at 5k/50M scale; savings scale linearly with the ceiling.}
\end{table}

\subsection*{Group G --- Z-loss and Router Stability}
\begin{table}[h]\centering
\begin{tabular}{lrrlrrrr}
\toprule
Run & $\alpha_z$ & clamp & batch & ppl & iters & savings & cv \\
\midrule
G0/A0 (baseline) & 1e-3 & 10.0 & 64 & 1342 & 3.00$\pm$0.01 & 62.5\% & 0.31 \\
G1 & 1e-2 & 10.0 & 64 & 1342 & 3.00$\pm$0.01 & 62.5\% & 0.48 \\
G2 & 1e-3 & off & 64 & 1342 & 2.00$\pm$0.00 & 75.0\% & 0.19 \\
G3 & 1e-3 & 10.0 & 128 & 1334 & 3.00$\pm$0.01 & 62.5\% & 0.19 \\
\bottomrule
\end{tabular}
\caption{Neither z-loss weight (1e-3 vs 1e-2) nor the logit clamp affects perplexity at 50M scale. A larger batch (G3) improves perplexity by 8 ppl; removing the logit clamp (G2) improves routing balance.}
\end{table}

\newpage
\section{Bootstrap Collapse: Resolution Attempts}
\label{app:collapse}

\begin{table}[h]\centering
\begin{tabular}{llll}
\toprule
Run & Configuration & Outcome & Collapse step \\
\midrule
III   & Geometric prior, no curriculum          & Collapsed to depth 1 & $\approx$300 \\
IV  & Geometric prior, slow $\alpha_{\mathrm{KL}}$ ramp      & Collapsed to depth 1 & 200--400 \\
V  & Curriculum release, geometric prior     & Collapsed to depth+1 & $\approx$900 \\
\textbf{II}  & Curriculum release, Gaussian prior, $\alpha_{\mathrm{KL}}=1.0$ & Stable through 30k steps & --- \\
\bottomrule
\end{tabular}
\caption{Empirical progression of depth-bootstrap collapse mitigation attempts. All runs use the 350M preset.}
\end{table}



\end{document}